\documentclass{sig-alt-release2}

\usepackage[colorlinks=true,linkcolor=blue,urlcolor=blue,citecolor=blue,anchorcolor=blue,hyperfigures]{hyperref}
\usepackage{float,subfig,epsfig}
\newcommand{\figwidth}{2.59in}

\begin{document}

\conferenceinfo{Proceedings of the Genetic and Evolutionary Computation Conference, GECCO'09,} {July 8--12, 2009, Montr\'{e}al, Qu\'{e}bec, Canada, pp. 1299--1306.}
\CopyrightYear{2009} 
\crdata{doi:10.1145/1569901.1570075} 
\clubpenalty=10000 
\widowpenalty = 10000

\title{Discrete Dynamical Genetic Programming in XCS} 

\numberofauthors{2} 

\author{
	\alignauthor
	Richard J. Preen\\
	\affaddr{Department of Computer Science}\\
	\affaddr{University of the West of England}\\
	\affaddr{Bristol, BS16 1QY, UK}\\
	\email{richard2.preen@uwe.ac.uk}
	\alignauthor
	Larry Bull\\
	\affaddr{Department of Computer Science}\\
	\affaddr{University of the West of England}\\
	\affaddr{Bristol, BS16 1QY, UK}\\
	\email{larry.bull@uwe.ac.uk}
}

\maketitle
\begin{abstract}
	A number of representation schemes have been presented for use within Learning Classifier Systems, ranging from binary encodings to neural networks. This paper presents results from an investigation into using a discrete dynamical system representation within the XCS Learning Classifier System. In particular, asynchronous random Boolean networks are used to represent the traditional condition-action production system rules. It is shown possible to use self-adaptive, open-ended evolution to design an ensemble of such discrete dynamical systems within XCS to solve a number of well-known test problems.  
\end{abstract}
\section{Introduction}

Traditionally, learning classifier systems (LCS)~\cite{Holland:1976} use a ternary encoding to generalise over the environmental inputs and to associate appropriate actions. A number of representations have previously been presented beyond this scheme however, including real numbers~\cite{Wilson:2000}, LISP S-expressions~\cite{LanziPerrucci:1999}, fuzzy logic~\cite{Valenzuela-Rendon:1991} and neural networks~\cite{Bull:2002}. To date, no temporally dynamic representation schemes have been used in LCS, a potentially important approach since temporal behaviour of such kinds is viewed as a significant aspect of cognition in general. 

In this paper we explore the use of a dynamical system representation within XCS~\cite{Wilson:1995}---what is herein termed ``dynamical genetic programming'' (DGP). Traditional tree-based genetic programming (GP)~\cite{Koza:1992} has been used within LCS both to calculate the action~\cite{AhluwaliaBull:1999} and to represent the condition~\cite{LanziPerrucci:1999}. DGP uses a graph-based representation, each node of which is constantly updated with asynchronous parallelism, and evolved using an open-ended, self-adaptive scheme. In the discrete case, each node is a Boolean function and therefore equivalent to a form of random Boolean network (RBN) (e.g.,~\cite{Kauffman:1993}). We show that XCS is able to solve a number of well-known immediate and delayed reward tasks using this temporally dynamic knowledge representation scheme.
\section{Related Work}

A number of representations have been presented by which to enable the evolution of computer programs, the most common being tree-based LISP S-expressions~\cite{LanziPerrucci:1999}. Other forms of GP include the use of machine code instructions (e.g.,~\cite{Banzhaf:1993}) and finite state machines (e.g.,~\cite{Fogel:1965}). Most relevant to the form of GP used in this paper is the small amount of prior work on graph-based representations. Teller and Veloso's ``neural programming''~\cite{TellerVeloso:1996} uses a directed graph of connected nodes, each with functionality defined in the standard GP way, with recursive connections included. Significantly, each node is executed with synchronous parallelism for some number of cycles before an output node's value is taken. Poli (e.g.,~\cite{PujolPoli:1998}) presented a very similar scheme wherein the graph is placed over a 2D grid and executes its nodes synchronously in parallel. Other examples of graph-based GP typically contain sequentially updating nodes (e.g.,~\cite{Miller:1999}). Schmidt and Lipson~\cite{SchmidtLipson:2007} have recently demonstrated a number of benefits from graph encodings over traditional trees, such as reduced bloat and increased computational efficiency.

As noted above, tree-based S-expressions have been used within LCS. Recently, Wilson~\cite{Wilson:2008} has explored the use of a form of gene expression programming (GEP)~\cite{Ferreira:2006} within LCS. Here the rules are represented as expression trees that are evaluated by assigning the environmental inputs to the tree's terminals, evaluating the tree, and then comparing the result with a predetermined threshold. Whenever the threshold value is exceeded, the rule is added to the match set.

The most common form of discrete dynamical system is the cellular automaton (CA)~\cite{VonNeumann:1966}, which consists of an array of cells (lattice of nodes) where the cells exist in states from a finite set and update their states with synchronous parallelism in discrete time. Traditionally, each cell calculates its next state depending upon its current state and the states of its closest neighbours. That is, CAs may be seen as a graph with a (typically) restricted topology. Packard~\cite{Packard:1988} was the first to use evolutionary computing techniques to design CAs such that they exhibit a given emergent global behaviour. Following Packard, Mitchell et~al. (e.g.,~\cite{Mitchell:1993}) have investigated the use of a genetic algorithm (GA)~\cite{Holland:1975} to learn the rules of uniform binary CAs. As in Packard's work, the GA produces the entries in the update table used by each cell, candidate solutions being evaluated with regard to their degree of success for the given task. Andre et~al.~\cite{Andre:1999} repeated Mitchell et~al.'s work whilst using traditional GP to evolve the update rules. They report similar results. Sipper (e.g.,~\cite{Sipper:1997}) presented a non-uniform, or heterogeneous, approach to evolving CAs. Each cell of a 1- or 2D CA is also viewed as a GA population member, mating only with its lattice neighbours and receiving an individual fitness. He shows an increase in performance over Mitchell et~al.'s work by exploiting the potential for spatial heterogeneity in the tasks. Sipper and Ruppin~\cite{SipperRuppin:1997} extended this approach to enable heterogeneity in the node connectivity, along with the node function; they evolved a form of random Boolean networks.

\section{Random Boolean Networks}

\begin{figure}[t]
	\centering
	\epsfig{file=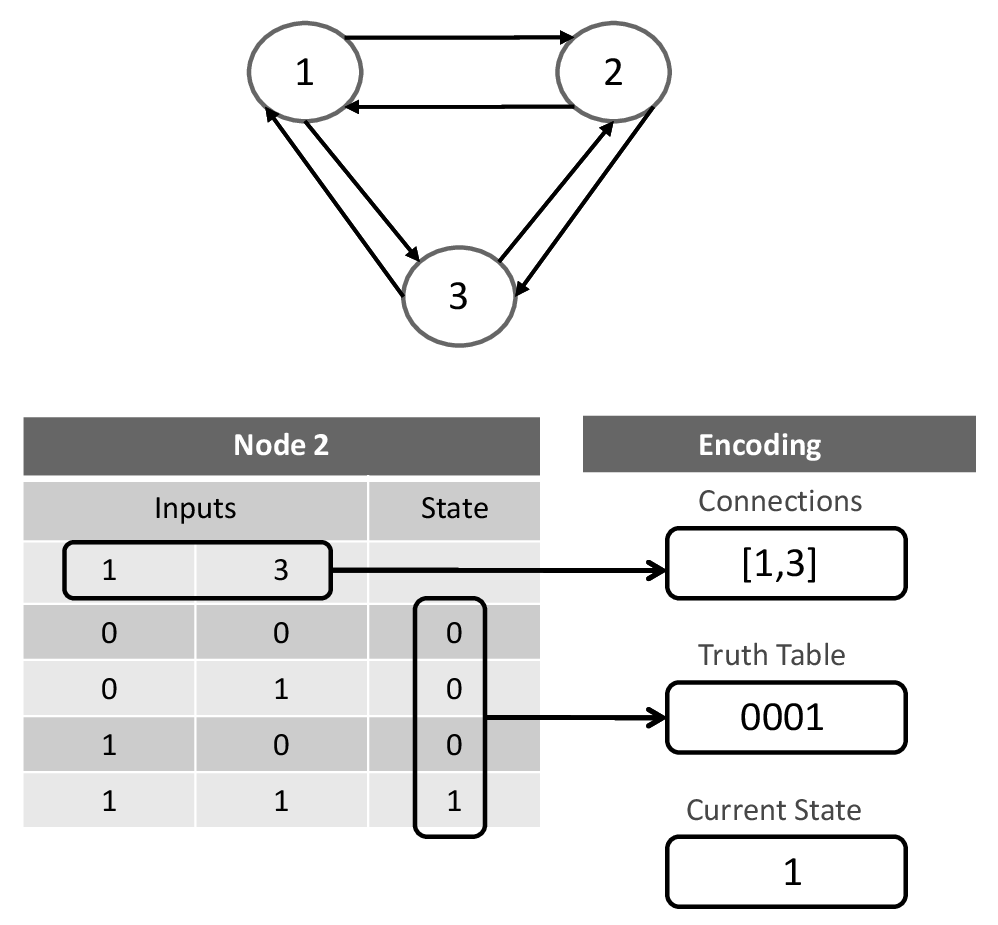,width=2.5in}
	\caption{Example Random Boolean Network and node encoding.}
	\label{fig:ExampleRBN}
\end{figure}

The discrete dynamical systems known as random Boolean networks (RBN) were originally introduced by Kauffman (see~\cite{Kauffman:1993}) to explore aspects of biological genetic regulatory networks. Since then they have been used as a tool in a wide range of areas, such as self-organisation (e.g.,~\cite{Kauffman:1993}) and computation (e.g.,~\cite{MesotTeuscher:2005}). An RBN typically consists of a network of $N$ nodes, each performing a Boolean function with $K$ inputs from other nodes in the network, all updating synchronously (see Figure~\ref{fig:ExampleRBN}). As such, RBN may be viewed as a generalisation of binary CAs. Since they have a finite number of possible states ($2^N$) and they use deterministic Boolean functions, the dynamics of RBN eventually fall into a basin of attraction. It is well-established that the value of $K$ affects the emergent behaviour of RBN wherein attractors typically contain an increasing number of states with increasing $K$. 3 phases of behaviour are suggested: ordered when $K=1$, with attractors consisting of 1 or a few states; chaotic when $K>3$, with a very large number of states per attractor; and, a critical regime around $K=2$, where similar states lie on trajectories that tend to neither diverge nor converge and 5--15\% of nodes change state per attractor cycle (see~\cite{Kauffman:1993} for discussions of this critical regime, e.g., with respect to perturbations). Analytical methods have been presented by which to determine the typical time taken to reach a basin of attraction and the number of states within such basins for a given degree of connectivity $K$ (see~\cite{Kauffman:1993}).

Closely akin to the work described here, Kauffman~\cite{Kauffman:1993} describes the use of simulated evolution to design RBN that must play a (mis)matching game wherein mutation is used to change connectivity, the Boolean functions, $K$ and $N$. He reports the typical emergence of high fitness solutions with $K$=2 to 3, together with an increase in $N$ over the initialised size. As noted above, traditional RBN consist of $N$ nodes updating synchronously in discrete time steps, but asynchronous versions have also been presented, after~\cite{HarveyBossomaier:1997}, leading to a classification of the space of possible forms of RBN~\cite{Gershenson:2002}. Asynchronous forms of CA have also been explored (e.g.,~\cite{IngersonBuvel:1984}) wherein it is often suggested that asynchrony is a more realistic underlying assumption for many natural and artificial systems.

Asynchronous logic devices are known to have the potential to consume less power and dissipate less heat~\cite{WernerAkella:1997}, which may be exploitable during efforts towards hardware implementations of such systems. Asynchronous logic is also known to have the potential for improved fault tolerance, particularly through delay insensitive schemes (e.g.,~\cite{DiLala:2007}). This may also prove beneficial for hardware implementations.

Harvey and Bossomaier~\cite{HarveyBossomaier:1997} showed that asynchronous RBN exhibit either point attractors, as seen in asynchronous CAs, or ``loose'' attractors where ``the network passes indefinitely through a subset of its possible states''~\cite{HarveyBossomaier:1997} (as opposed to distinct cycles in the synchronous case). Thus the use of asynchrony represents another feature of RBN with the potential to significantly alter their underlying dynamics thereby offering another mechanism by which to aid the simulated evolutionary design process for a given task. Di~Paolo~\cite{DiPaolo:2001} showed it is possible to evolve asynchronous RBN that exhibit rhythmic behaviour at equilibrium. Asynchronous CAs have also been evolved (e.g.,~\cite{SipperRuppin:1997}).

\section{Discrete DGP-XCS}

\begin{figure}[t]
	\centering
	\epsfig{file=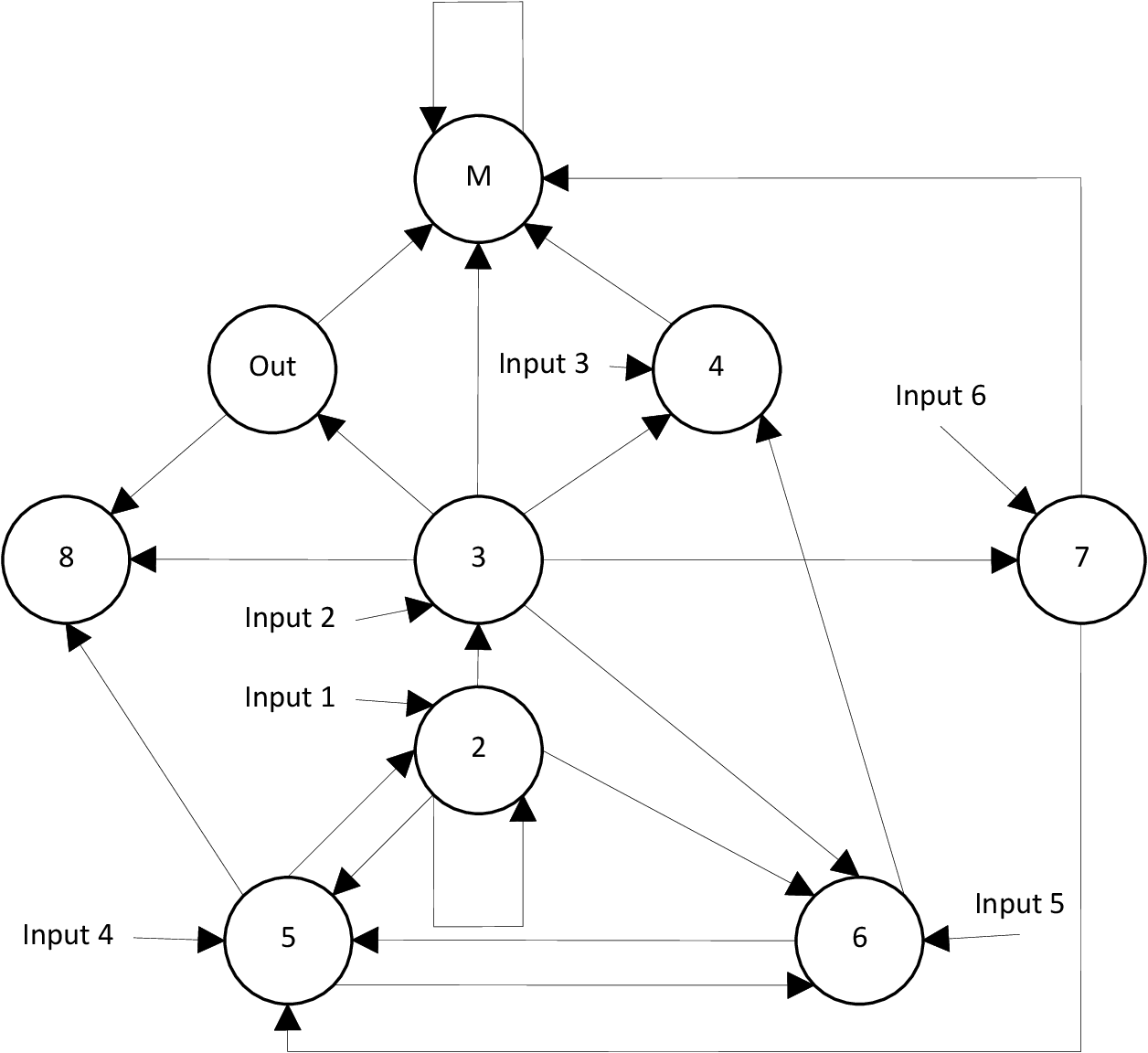,width=3.3in}
	\vspace{0.2in}

	\begingroup
	\fontsize{6pt}{6pt}\selectfont
	\begin{tabular}{l l l}
		\multicolumn{3}{l}{Prediction 1000. Error: 0.0. Accuracy: 1.0.}\\
		\multicolumn{3}{l}{Experience: 822. GASetSize: 70.1. GATimeStamp: 99947}\\ \\
		& Truth Table: & Connections: \\
		Node 0 (M): & 10011000100000001110011010101000 & 7, 4, 0, 3, 1 \\
		Node 1 (out): & 10 & 3 \\
		Node 2 (I): & 00011111 & {\em Input1}, 2, 5 \\
		Node 3 (I): & 0001 & {\em Input2}, 2 \\
		Node 4 (I): & 11101110 & {\em Input3}, 6, 3 \\
		Node 5 (I): & 0110110100001010 & {\em Input4}, 2, 7, 6 \\
		Node 6 (I): & 0001011101010101 & {\em Input5}, 5, 2, 3 \\
		Node 7 (I): & 0100 & {\em Input6}, 3 \\
		Node 8 (N): & 00010111 & 3, 1, 5 \\
	\end{tabular}
	\endgroup

	\caption{An evolved dDGP-XCS 6-bit multiplexer asynchronous rule.}
	\label{fig:dDGP-examplerule}
\end{figure}

To use asynchronous RBN as the rules within XCS, the following scheme is adopted. Each of an initial randomly created rule's nodes has $K$ randomly assigned connections, here $1 \leq K \leq 5$. There are as many nodes $N$ as input fields $I$ for the given task and its outputs $O$, plus 1 other, as will be described, i.e., $N=I+O+1$. The first connection of each input node is set to the corresponding locus of the input message. The other connections are assigned at random within the RBN as usual. In this way, the current input state is always considered along with the current state of the RBN itself per network update cycle by such nodes (see Figure~\ref{fig:dDGP-examplerule}). Nodes are initialised randomly each time the network is run to determine [M], etc. The population is initially empty and covering is applied to generate rules as in the standard XCS approach.

Matching consists of executing each rule for $T$ cycles based on the current input. The value of $T$ is chosen to be a value typically within the basin of attraction of the RBN. Asynchrony is here implemented as a randomly chosen node being updated on a given cycle, with as many updates per overall network update cycle as there are nodes in the network before an equivalent cycle to 1 in the synchronous case is said to have occurred. See~\cite{Gershenson:2002} for alternative schemes.

In this study, when well-known Boolean problems are explored there are only 2 possible actions and thus only 1 output node is required. Where well-known maze problems are explored there are 8 possible actions and accordingly 3 required output nodes. An extra ``matching'' node is also required to enable RBNs to (potentially) only match specific sets of inputs. If a given RBN has a logical `0' on the match node, regardless of its output node's state, the rule does not join [M] (see Figure~\ref{fig:dDGP-examplerule}). This scheme has also been exploited within neural LCS~\cite{Bull:2002}. A `windowed approach' is utilised where the output is decided by the most common state over the last $W$ steps up to $T$. For example, if the last few states on a node updating prior to cycle $T$ is 0101001 and $W=3$, then the ending node's state would be `0' and not `1'. In this paper, $W$ is set to 3. Thereafter, match set and action set processing proceeds as standard in XCS (the reader is referred to~\cite{ButzWilson:2001} for an algorithmic description of XCS).

When covering is necessitated, a randomly constructed RBN is created and then executed for $T$ cycles to determine the status of the match and output nodes. This procedure is repeated until an RBN is created that matches the environment state. 

Parameter self-adaptation was first explored in LCS by Bull et~al.~\cite{Bull:2000} wherein the mutation rate is a locally evolving entity in itself; each rule has its own mutation rate $\mu$ Mutation only is used here and applied to the node's truth table and connectivity map at rate $\mu$. A node's truth table is represented by a binary string and its connectivity by a list of $K$ integers in the range $[1, N]$. Since each node has a given fixed $K$ value, each node maintains a binary string of length $2^K$, which forms the entries in the look-up table for each of the possible $2^K$ input states of that node, i.e., as in the aforementioned work of Packard~\cite{Packard:1988} on evolving CAs, for example. These strings are subjected to mutation on reproduction at the self-adapting rate $\mu$ for that rule. Hence, within the RBN representation, evolution can define different Boolean functions for each node within a given network rule, along with its connectivity map. Specifically, each rule has its own mutation rate stored as a real number and initially seeded uniform randomly in the range $[0,1]$. This parameter is passed to its offspring. The offspring then applies its mutation rate to itself using a Gaussian distribution, i.e., $\mu' = \mu e^{N(0,1)}$, before mutating the rest of the rule at the resulting rate.

Due to the need for a possible different number of nodes within the rules for a given task, the DGP scheme is also of variable length. Once the truth table and connections have been mutated, a new randomly connected node is either added or the last added node is removed with the same probability $\mu$. The latter case only occurs if the network currently consists of more than the initial number of nodes. Thus DGP is temporally dynamic both in the search process and the representation scheme. Evolving variable-length solutions via mutation only has previously been explored a number of times, e.g.,~\cite{Fogel:1965}. Traditional GP can be seen to primarily rely upon recombination to search the space of possible tree sizes, although the standard mutation operator effectively increases or decreases tree size also. Whenever an offspring classifier is created and no changes occur to its RBN when undergoing mutation, the parent's numerosity is increased and mutation rate set to the offspring's.

\section{Experimentation}

\subsection{Multiplexer}

We now apply this discrete version of DGP-XCS (dDGP-XCS) to the well-known multiplexer task. These Boolean functions are defined for binary strings of length $l = x + 2^x$ under which the $x$ bits index into the remaining $2^x$ bits, returning the value of the indexed bit. The correct classification to a randomly generated input results in a payoff of 1000, otherwise 0.

\begin{figure}[t]
	\centering
	\subfloat[Performance (circle), error (square), macro-classifiers (triangle) and mutation rate (diamond).]
	{ \label{fig:dDGP-XCS-6MUX-Asynch} \epsfig{file=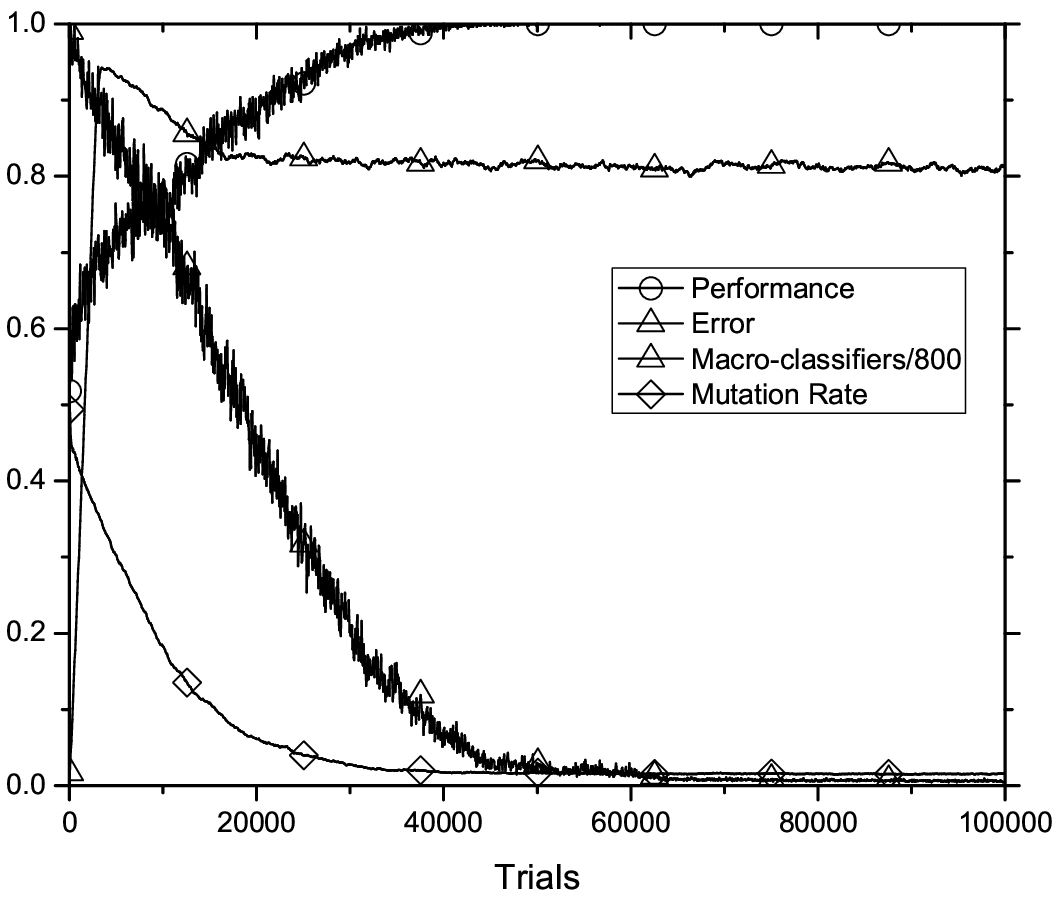,width=\figwidth} } \hspace{1in}
	\subfloat[Average number of nodes (circle) and average number of connections (square).] 
	{ \label{fig:dDGP-XCS-6MUX-Asynch-Top} \epsfig{file=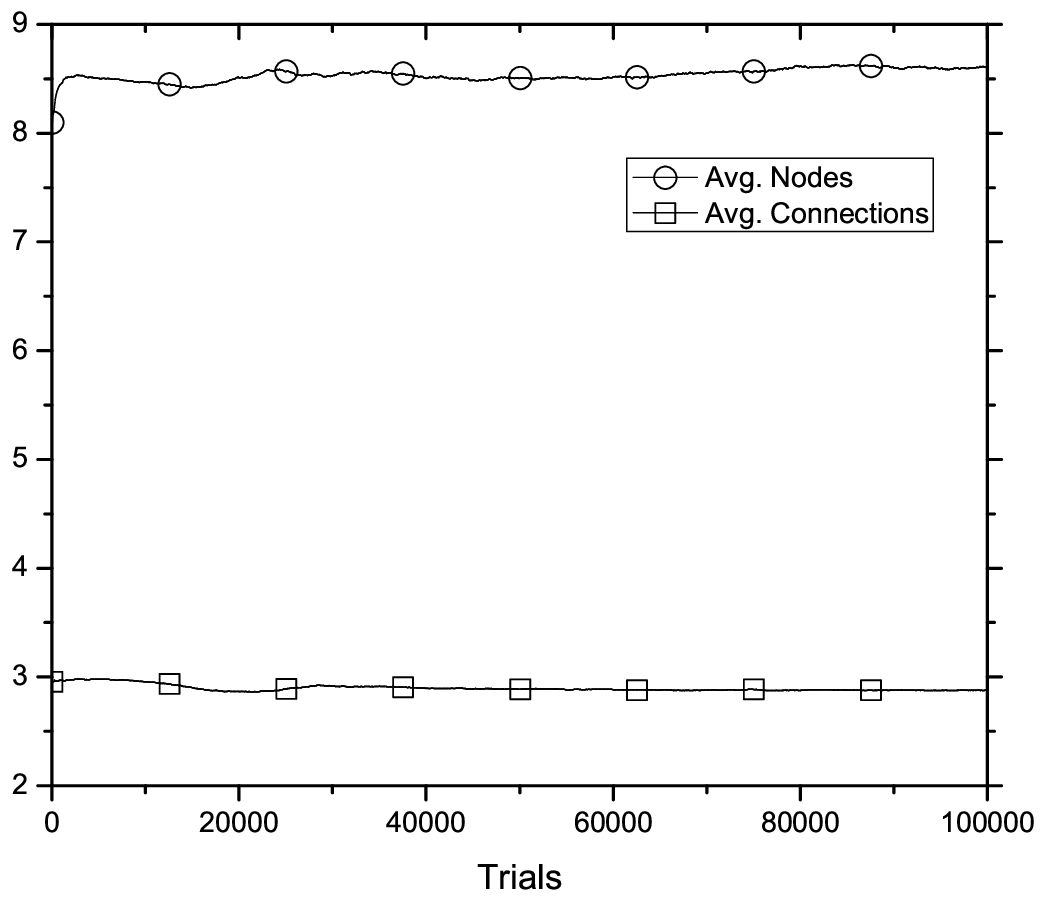,width=\figwidth} }
	\caption{dDGP-XCS 6-bit Multiplexer Performance}        
	\label{fig:dDGP-XCS-6MUX}
\end{figure}

Figure~\ref{fig:dDGP-XCS-6MUX} shows the performance of the constructed system on the 6-bit multiplexer problem updated asynchronously with $P=800$, $\nu=5$, $\theta_{GA}=25$, $\beta=0.2$, $p_{expl}=1.0$, $T=25$, $W=3$, and $N_{init} = 8$ (6 inputs, 1 output, 1 match node). After Wilson~\cite{Wilson:1995}, performance from exploit trials only is recorded (fraction of correct responses are shown), using a 50-point running average, averaged over 10 runs.

\begin{figure}[t]
	\centering
	\epsfig{file=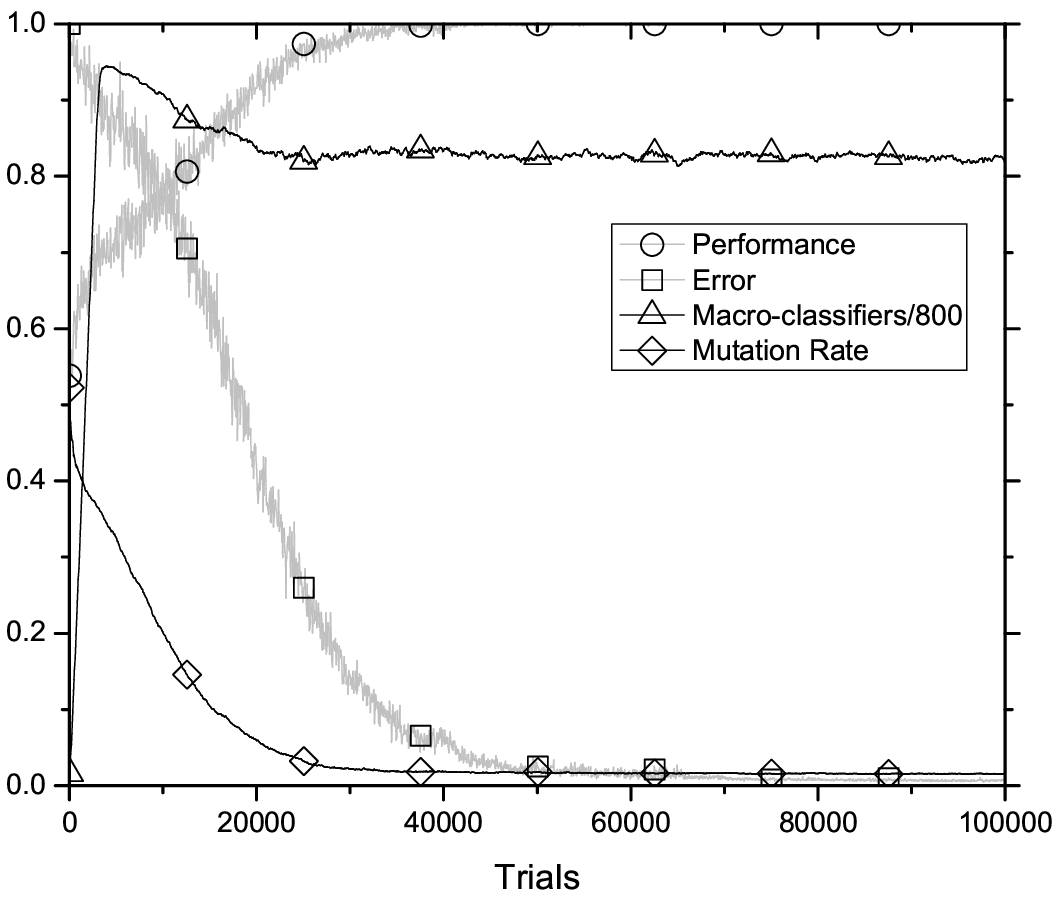,width=\figwidth}
	\caption{dDGP-XCS 6-bit Multiplexer synchronous performance (circle), error (square), macro-classifiers (triangle) and mutation rate (diamond).}
	\label{fig:dDGP-XCS-6MUX-synch}
\end{figure}

From Figure~\ref{fig:dDGP-XCS-6MUX-Asynch} it can be seen that a near optimal solution is learnt around 35,000 trials and optimality is observed around trial 58,000. The parameter governing RBN mutation (see Figure~\ref{fig:dDGP-XCS-6MUX-Asynch}) declines rapidly until reaching a bottom around 40,000 trials, which is shortly after discovering an optimal solution. The number of (non-unique) rules initially grows rapidly, before declining to around 650. Furthermore, the average degree of connectivity $K$ decreases fractionally, whilst, on average, each network grows approximately 1 extra node (see Figure~\ref{fig:dDGP-XCS-6MUX-Asynch-Top}. This behaviour indicates that the evolutionary process is able to identify an appropriate typical topology with which to generate complex behaviour, i.e., in this case a computation. For other tasks, other values of $K$ may prove beneficial; high $K$ may be expected in random number generation, for example. It can be noted that a growth event under which a new node is added into an RBN is essentially neutral here since the new node receives inputs from the existing nodes (or itself) on addition but only provides inputs to other nodes after subsequent connectivity mutations. For comparative purposes, Figure~\ref{fig:dDGP-XCS-6MUX-synch} shows the performance with the same parameters on the 6-bit multiplexer when updated synchronously. It is shown that the performance is very similar regardless of the updating scheme and that there is thus apparently very little overhead when updating asynchronously, with the possible benefits mentioned above.
Figure~\ref{fig:dDGP-examplerule} provides an illustration of a rule generated whilst solving the 6-bit multiplexer problem when updated asynchronously. There is 1 new node in addition to the initial 8. The truth table shows to which state each node will transition, given each of the possible inputs. For example, the output node (node 1) has a truth table of `10', which is synonymous with a NOT gate where if node 3 is in state `0' then the output node will be set to `1', and if node 3 is in state `1' then the output node will be set to `0'. The truth table of node 3 is synonymous with an AND gate, etc.

The rule has a prediction of 1000 and an Error of 0, whilst having an experience of 822, showing that this is a highly accurate rule. Analysis of this RBN rule was undertaken by executing it for each of the 64 6-bit inputs. Each input was run 20 times with $T=25$ and $W=3$. The results show that for the majority of environment states the network will return a false match node, preventing it from being added to [M]. However, the network is general as the match node will always return true when the environment states are 110000, 110010, 110100, 110110, 111000, 111010, 111100, and 111110. In all of those cases the output node always advocates action `0'. In addition, there are several environment states for which the match node will only sometimes return true. However, in all cases when the match node does permit the rule to be added to [M], the action advocated will always be consistent. There are 4 such additional environment states (010000, 010010, 011000, and 011010) for which the rule will match, albeit with a probability less than 50\%.

The rule in Figure~\ref{fig:dDGP-examplerule} was then re-run as before, however using a traditional synchronous updating scheme. The results of the match node and output nodes are extremely similar regardless of the updating mechanism. That is, XCS has evolved an RBN that is very robust to the random nature of the asynchronous updating, meaning it is accurate even for the relatively rare case of all nodes updating concurrently, i.e., the synchronous case.

\subsection{Maze Environments}

In addition to the single-step multiplexer problems, dDGP-XCS is applied to versions of 3 well-known multi-step maze environments, Woods~1 (see Figure~\ref{fig:woods1}), Maze~4 (see Figure~\ref{fig:maze4}), and Woods~101 (see Figure~\ref{fig:woods101}).

\begin{figure}[!tbh]
	\centering
	\subfloat[Number of Steps to Goal (circle).]
	{ \label{fig:dDGP-XCS-Woods1} \epsfig{file=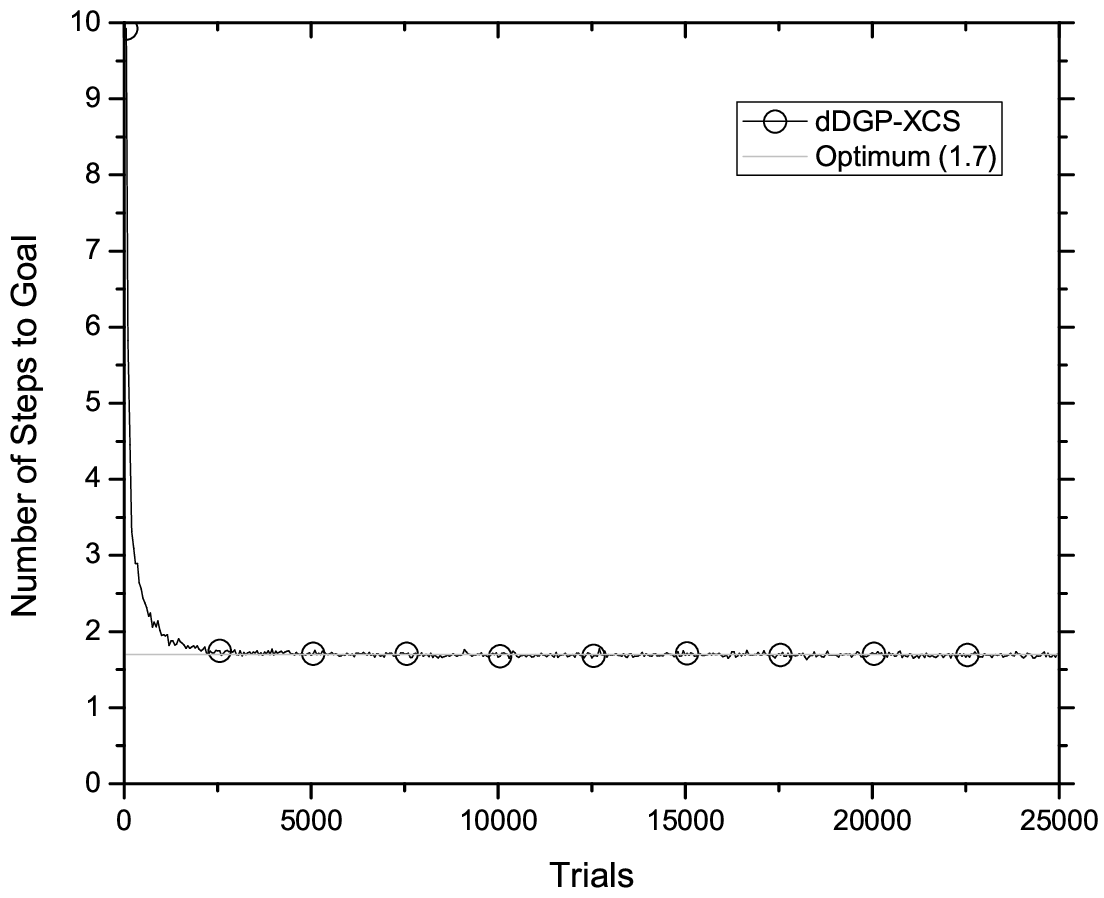,width=\figwidth} }
	\\
	\subfloat[Average mutation rate (square) and number of macro-classifiers (circle).]
	{ \label{fig:dDGP-XCS-Woods1-SizeMut} \epsfig{file=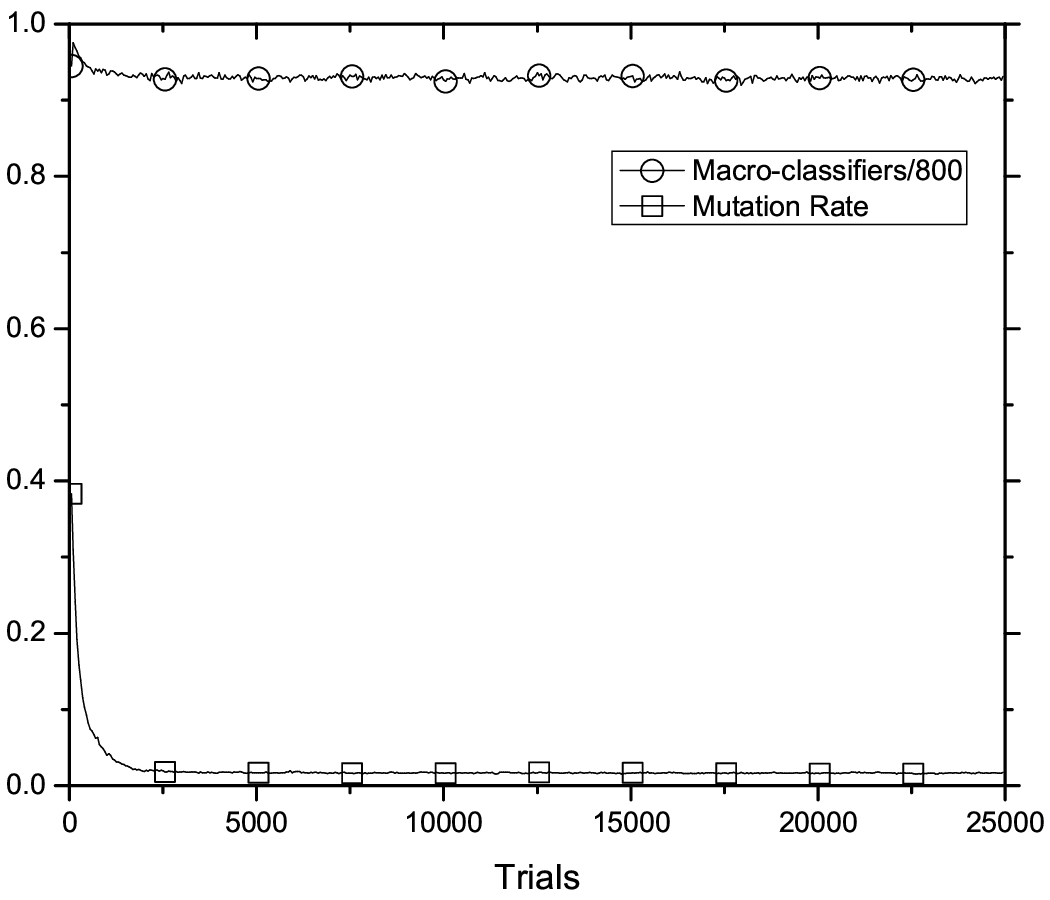,width=\figwidth} }\\
	\subfloat[Average number of nodes (circle) and average number of connections (square).]
	{ \label{fig:dDGP-XCS-Woods1-Topology} \epsfig{file=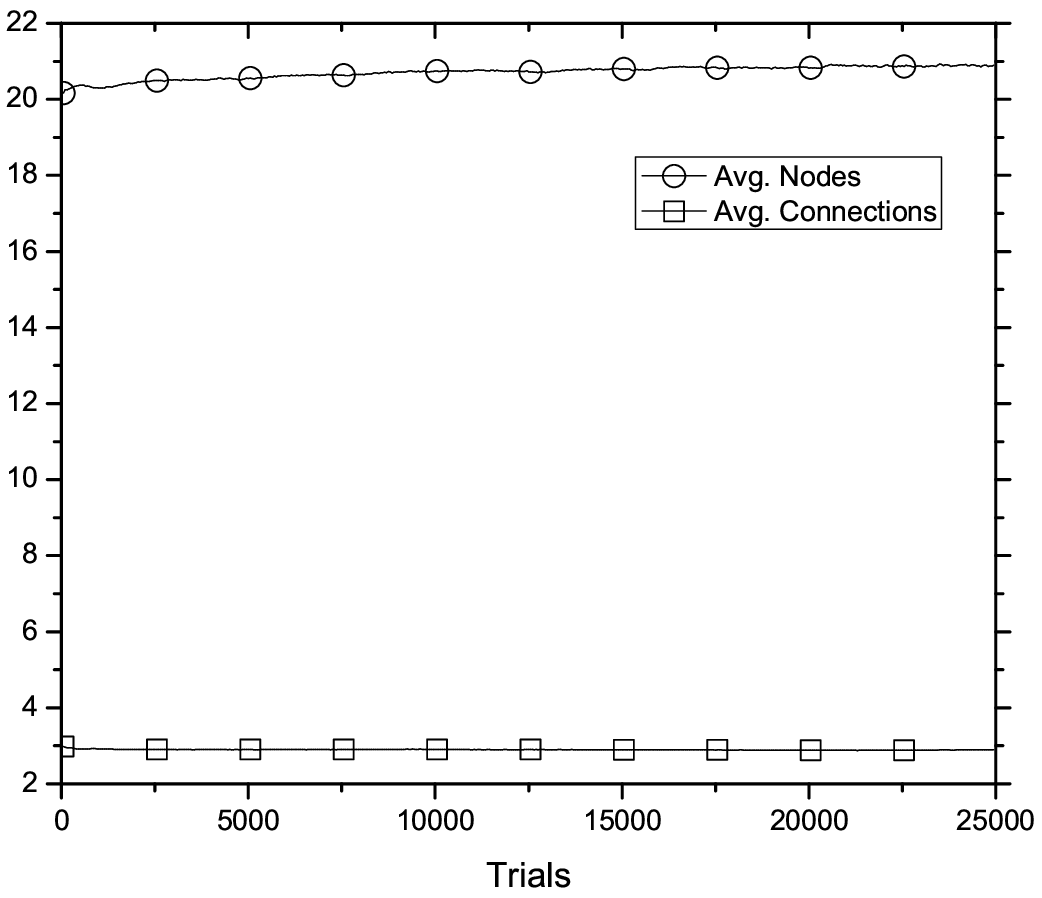,width=\figwidth} }
	\caption{dDGP-XCS Woods~1 Performance}
\end{figure}

\begin{figure}[!tbh]
	\centering
	\subfloat[Number of Steps to Goal (circle).]
	{ \label{fig:dDGP-XCS-Maze4-Perf} \epsfig{file=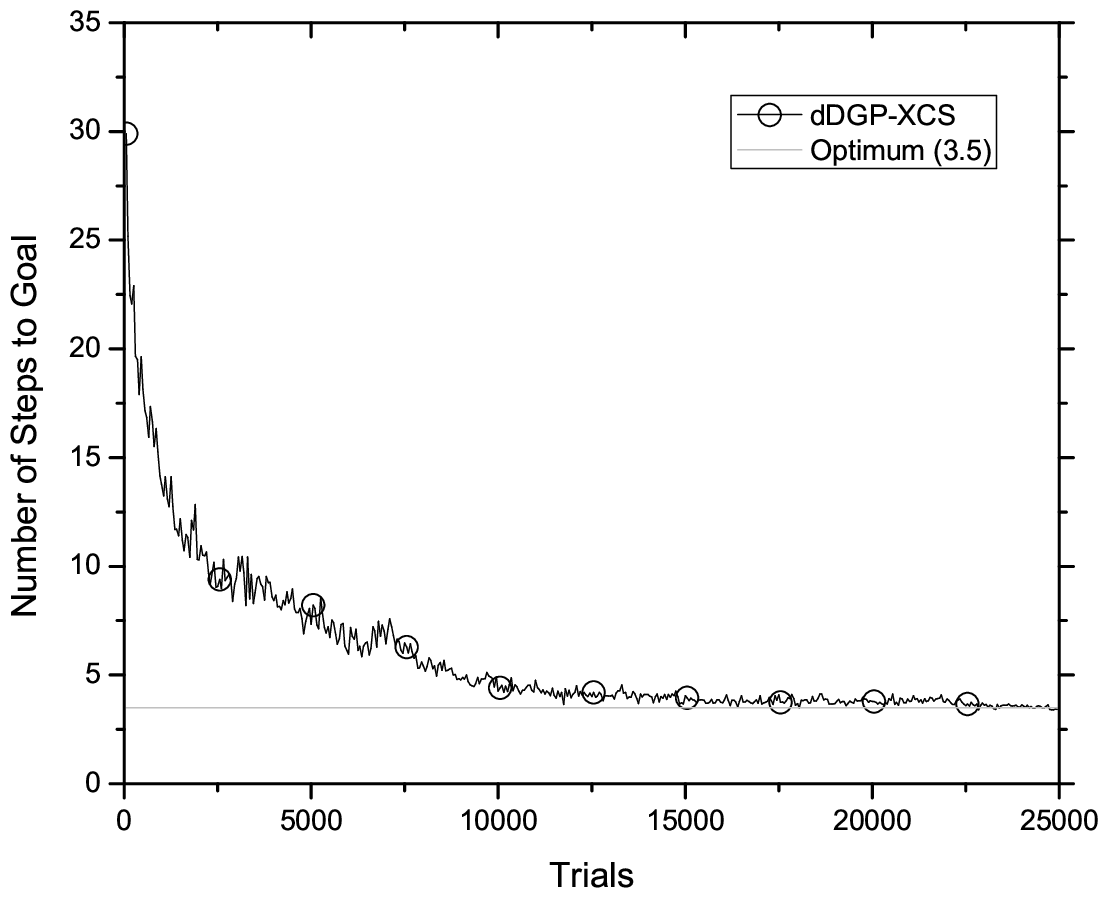,width=\figwidth} }
	\\
	\subfloat[Average mutation rate (square) and number of macro-classifiers (circle).]
	{ \label{fig:dDGP-XCS-Maze4-SizeMut} \epsfig{file=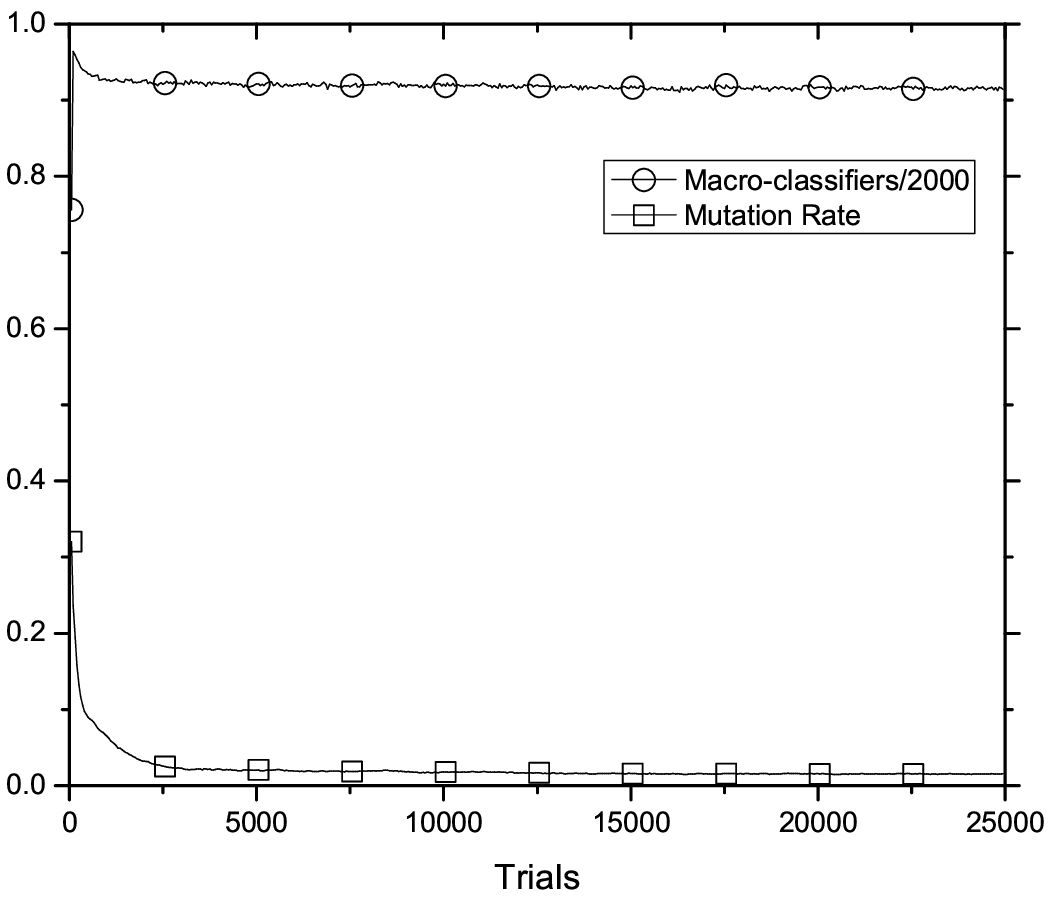,width=\figwidth} }\\
	\subfloat[Average number of nodes (circle) and average number of connections (square).]
	{ \label{fig:dDGP-XCS-Maze4-Topology} \epsfig{file=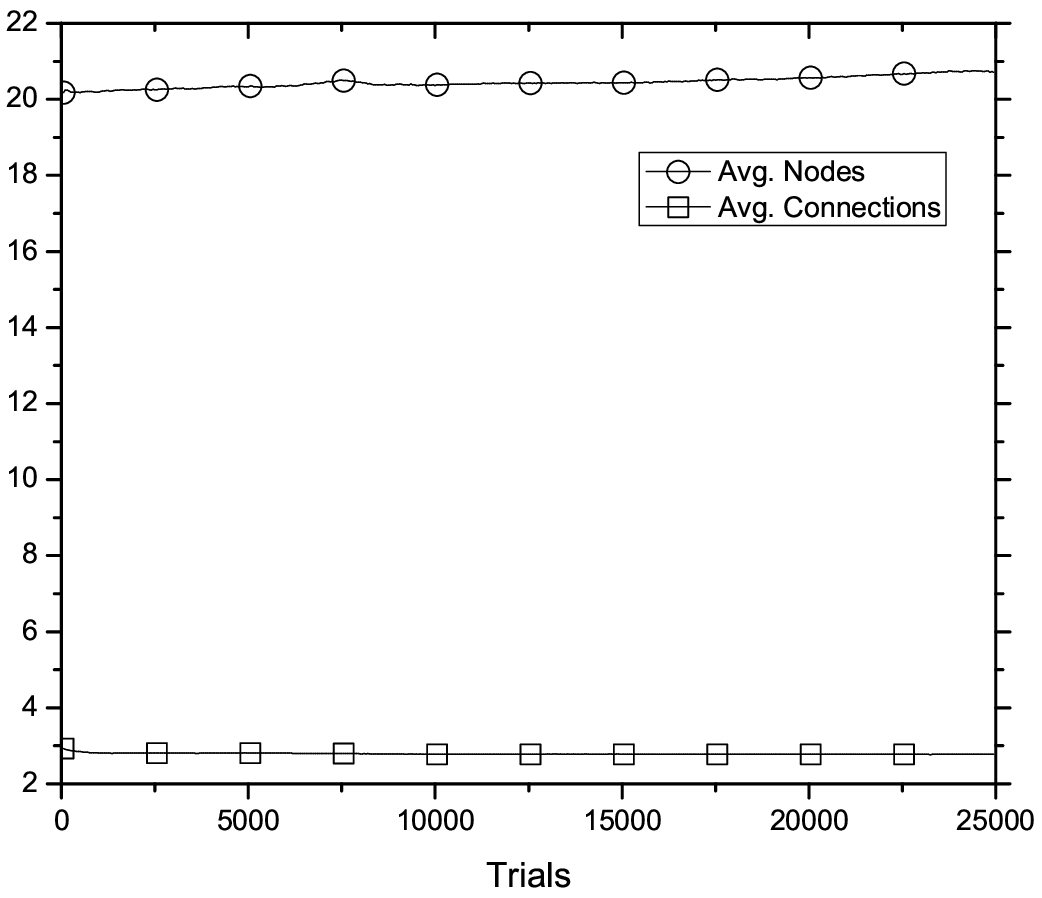,width=\figwidth} }
	\caption{dDGP-XCS Maze~4 Performance}
\end{figure}

Each cell in the maze environments is encoded with 2 binary bits, where white space is represented as a `*', obstacles as `O', and food as `F'. Furthermore, actions are encoded in binary as shown in Figure~\ref{fig:maze-encoding}. The task is simply to find the shortest path to the food (F) given a random start point. Obstacles (O) represent cells that cannot be occupied. A teletransportation mechanism is employed whereby a trial is reset if the agent has not reached the goal state within 50 discrete movements. In Woods~1 the optimal number of steps to the food is 1.7, in Maze~4 optimal is 3.5 steps, and in Woods~101 it is 2.9.
\begin{figure}[t]
	\centering
	\subfloat[Woods~1] { 
		\centering
		\begin{tabular}{ | c | c | c | c | c | }
			\hline
			* & * & * & * & * \\ \hline
			* & * & * & * & * \\ \hline
			O & O & F & * & * \\ \hline
			O & O & O & * & * \\ \hline
			O & O & O & * & * \\
			\hline
		\end{tabular}
	\label{fig:woods1} }%
	\subfloat[Maze~4] { 
		\centering
		\begin{tabular}{ | c | c | c | c | c | c | c |}
			\hline
			O & O & O & O & O & O & O \\ \hline
			O & * & * & * & * & * & O \\ \hline
			O & * & O & * & O & * & O \\ \hline
			O & * & O & F & O & * & O \\ \hline
			O & O & O & O & O & O & O \\
			\hline
		\end{tabular}
	\label{fig:maze4} }
	\\
	\subfloat[Woods~101] { 
		\centering
		\begin{tabular}{ |*{7}{@{\hspace{1mm}}c@{\hspace{1mm}}| } }
			\hline
			O & O & O & O & O & O & O \\ \hline
			O & * & * & * & * & * & O \\ \hline
			O & * & O & * & O & * & O \\ \hline
			O & * & O & F & O & * & O \\ \hline
			O & O & O & O & O & O & O \\
			\hline
		\end{tabular}
	\label{fig:woods101} } \\
	\subfloat[Maze Encoding.] {
		\centering
		\begin{tabular}{ | c | c | c@{\hspace{1mm}} | c | c | c |}
			\cline{1-2} \cline{4-6}
			Cell & Binary & & \multicolumn{3}{| c |}{Actions} \\ \cline{1-2} \cline{4-6}
			* & 00 & & 111 & 000 & 001 \\ \cline{1-2} \cline{4-6}
			O & 01 & & 110 & & 010 \\ \cline{1-2} \cline{4-6}
			F & 11 & & 101 & 100 & 011 \\ \cline{1-2} \cline{4-6}
		\end{tabular}
	\label{fig:maze-encoding} }
	\caption{Experimental Maze Environments and Encoding.}
	\label{fig:mazes}
\end{figure}
Figures~\ref{fig:dDGP-XCS-Woods1}--\ref{fig:dDGP-XCS-Woods1-Topology} show the performance of dDGP-XCS in the Woods~1 environment. The parameters used are identical to those applied in the aforementioned multiplexer experiments, except that $N_{init} = 20$ (16 inputs, 3 outputs, 1 match node) ($P=800$). As can be seen from Figure~\ref{fig:dDGP-XCS-Woods1}, optimality is observed around 2,500 trials. This roughly matches the performance of neural XCS using self-adaptive constructivism ($\approx$2,500 trials, $P=2000$)~\cite{Howard:2008} and faster than XCS using messy conditions ($\approx$8,000 trials, $P=800$)~\cite{Lanzi:1999b}, XCS using stack-based GP conditions ($\approx$10,000 trials, $P=1000$)~\cite{Lanzi:2003}, and XCS with LISP S-expression conditions ($\approx$5,000 trials, $P=800$)~\cite{LanziPerrucci:1999}. Figure~\ref{fig:dDGP-XCS-Woods1-SizeMut} shows that there is an average of 745 (non-unique) rules evolved. In addition, Figure~\ref{fig:dDGP-XCS-Woods1-SizeMut} shows that the mutation rate declines rapidly by 2,800 trials, shortly after the optimal solution is learnt. Figure~\ref{fig:dDGP-XCS-Woods1-Topology} shows that on average the networks add 1 extra node (from the original 20) and the average number of connections decreases slightly.
Figures~\ref{fig:dDGP-XCS-Maze4-Perf}--\ref{fig:dDGP-XCS-Maze4-Topology} present the performance of dDGP-XCS in the Maze~4 environment. The parameters used are identical to those in the Woods~1 environment, however a bigger population limit of $P=2000$ is used, reflecting the larger search space. Optimality is observed around trial 23,000 (see Figure~\ref{fig:dDGP-XCS-Maze4-Perf}), which is again similar to the performance observed using a neural XCS with self-adaptive constructivism ($\approx$23,000 trials, $P=3000$)~\cite{Howard:2008}. The average number of rules evolved is around 1,800 (see Figure~\ref{fig:dDGP-XCS-Maze4-SizeMut}). The average number of nodes in the networks also increases by almost 1, and the average number of connections declines slightly from 3 (see Figure~\ref{fig:dDGP-XCS-Maze4-Topology}). The parameter governing RBN mutation (Figure~\ref{fig:dDGP-XCS-Maze4-SizeMut}) declines rapidly after 4,000 trials, before finally stabilising after 15,000 trials.

The Woods~101 maze is a non-Markov environment containing 2 {\em communicating aliasing states}, i.e., 2 positions that border on the same non-aliasing state and are identically sensed, but require different optimal actions. Thus, to solve this maze optimally, a form of memory must be utilised (with at least 2 internal states). Optimal performance has previously been achieved in Woods~101 through the addition of a memory register mechanism in XCS~\cite{LanziWilson:2000}, a corporate XCS using rule-linkage~\cite{Tomlinson:2001}, and a neural LCS using recurrent links~\cite{BullHurst:2003}. Furthermore, in a proof of concept experiment, the cyclical directed graph from neural programming has been shown capable of representing rules with memory to solve Woods~101, however it was only found to do so twice in 50 experiments~\cite{BalanLuke:2004}.

The simplest form of short-term memory is a fixed-length buffer containing the $n$ most recent inputs; a common extension is to then apply a kernel function to the buffer to enable non-uniform sampling of the past values, e.g., an exponential decay of older inputs~\cite{Mozer:1994}. Simple forms of memory are {\em static}, i.e., the memory parameters are fixed in advance and the memory state is thus a predetermined function of the input sequence. However, it is not clear that biological systems make use of such shift registers. Registers require some interface with the environment that buffers the input so that it can be presented simultaneously. They impose a rigid limit on the duration of patterns, defining the longest possible pattern and requiring that all input vectors be of the same length. Furthermore, such approaches struggle to distinguish relative temporal position from absolute temporal position~\cite{Elman:1990}.

The hypothesis of inherent content-addressable memory existing within synchronous RBN due to different possible routes to a basin of attraction~\cite{Wuensche:2004} for the asynchronous case is here explored and extended by simply not resetting the node states on each step. A significant advantage of this approach is that each rule/network's short-term memory is variable-length and {\em adaptive}, i.e., the networks can adjust the memory parameters, selecting within the limits of the capacity of the memory, what aspects of the input sequence are available for computing predictions~\cite{Mozer:1994}. In addition, as open-ended evolution is used, the maximum size of the short-term memory is potentially also open-ended, increasing as the number of nodes within the network grows.

Here, nodes are initialised at random for the initial random placing in the maze but thereafter they are not reset for each subsequent matching cycle. Consequently, each network processes the environmental input and the final node states then become the starting point for the next processing cycle, whereupon the network receives the new environmental input and places the network on a trajectory toward a (potentially) different locally stable limit point. A network given the same environmental input (i.e., the agent's current maze perception) but with different initial node states (representing the agent's history through the maze) may fall into a different basin of attraction (advocating a different action). {\em Thus the rules' dynamics are (potentially) constantly affected by the inputs as the system executes.}

\begin{figure}[!tbh]
	\centering
	\subfloat[Number of Steps to Goal (circle).]
	{ \label{fig:dDGP-XCS-Woods101} \epsfig{file=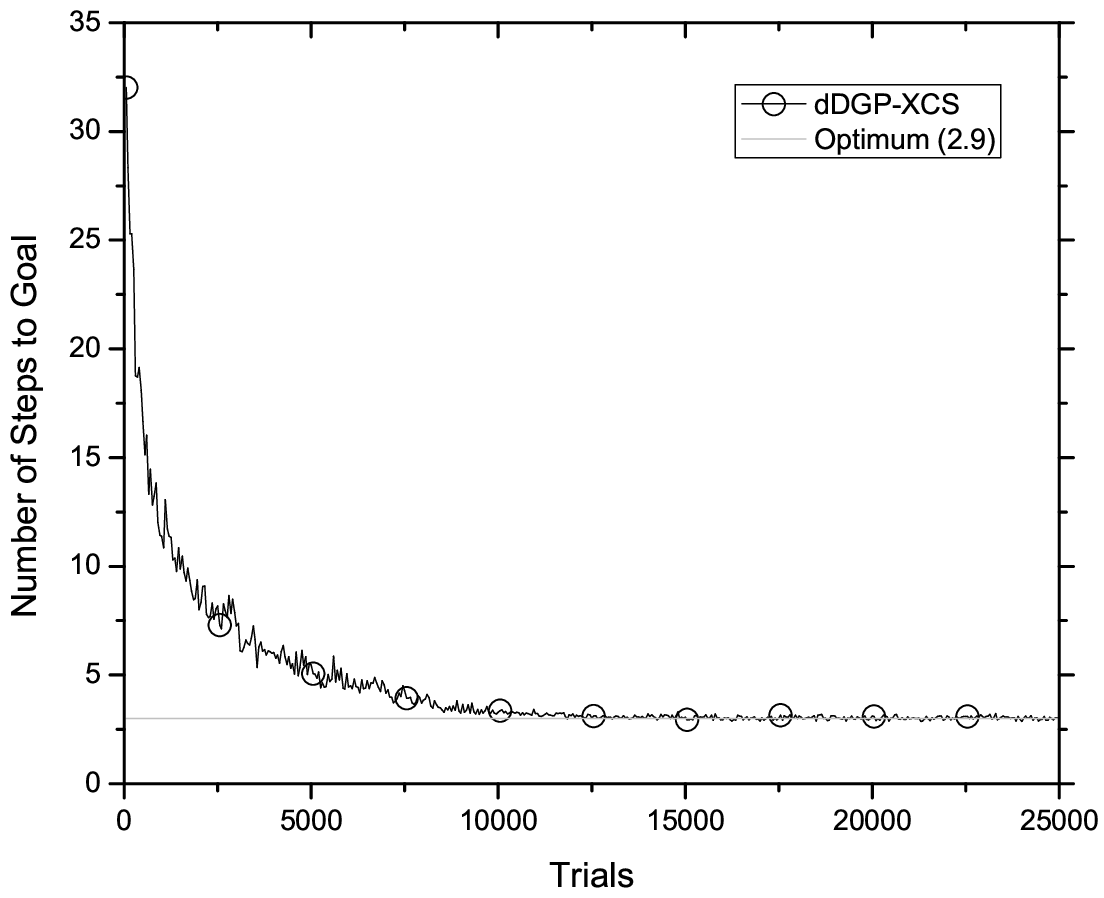,width=\figwidth} }\\
	\subfloat[Average mutation rate (square) and number of macro-classifiers (circle).]
	{ \label{fig:dDGP-XCS-Woods101-SizeMut} \epsfig{file=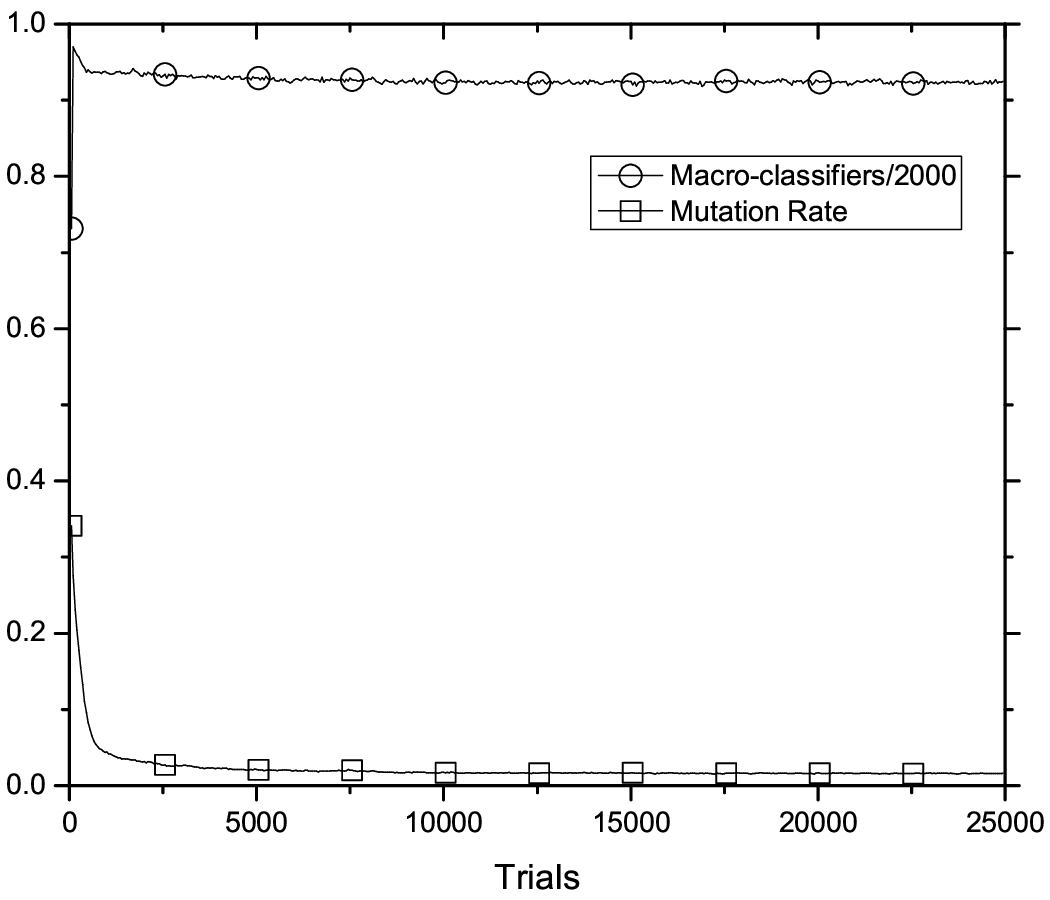,width=\figwidth} }
	\\
	\subfloat[Average number of nodes (circle) and average number of connections (square).]
	{ \label{fig:dDGP-XCS-Woods101-Topology} \epsfig{file=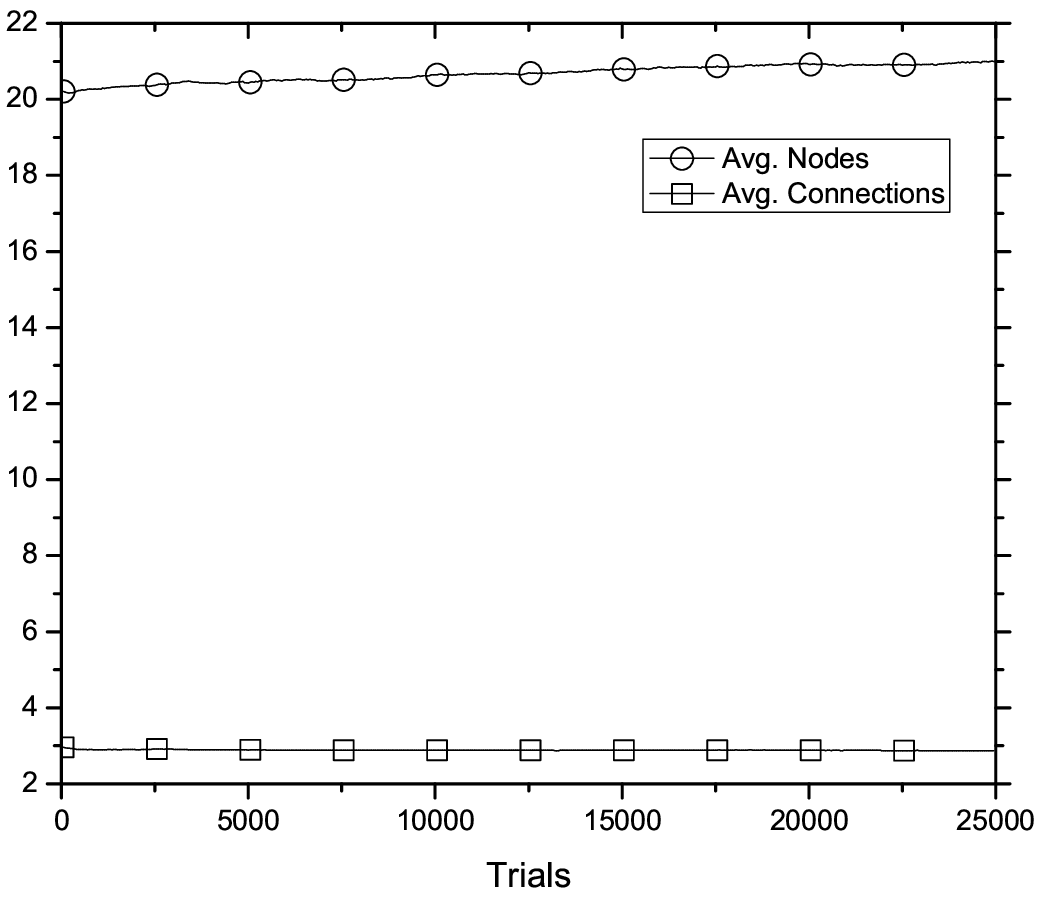,width=\figwidth} }
	\caption{dDGP-XCS Woods~101 Performance}
\end{figure}
\begin{figure}[!tbh]
	\centering
	\epsfig{file=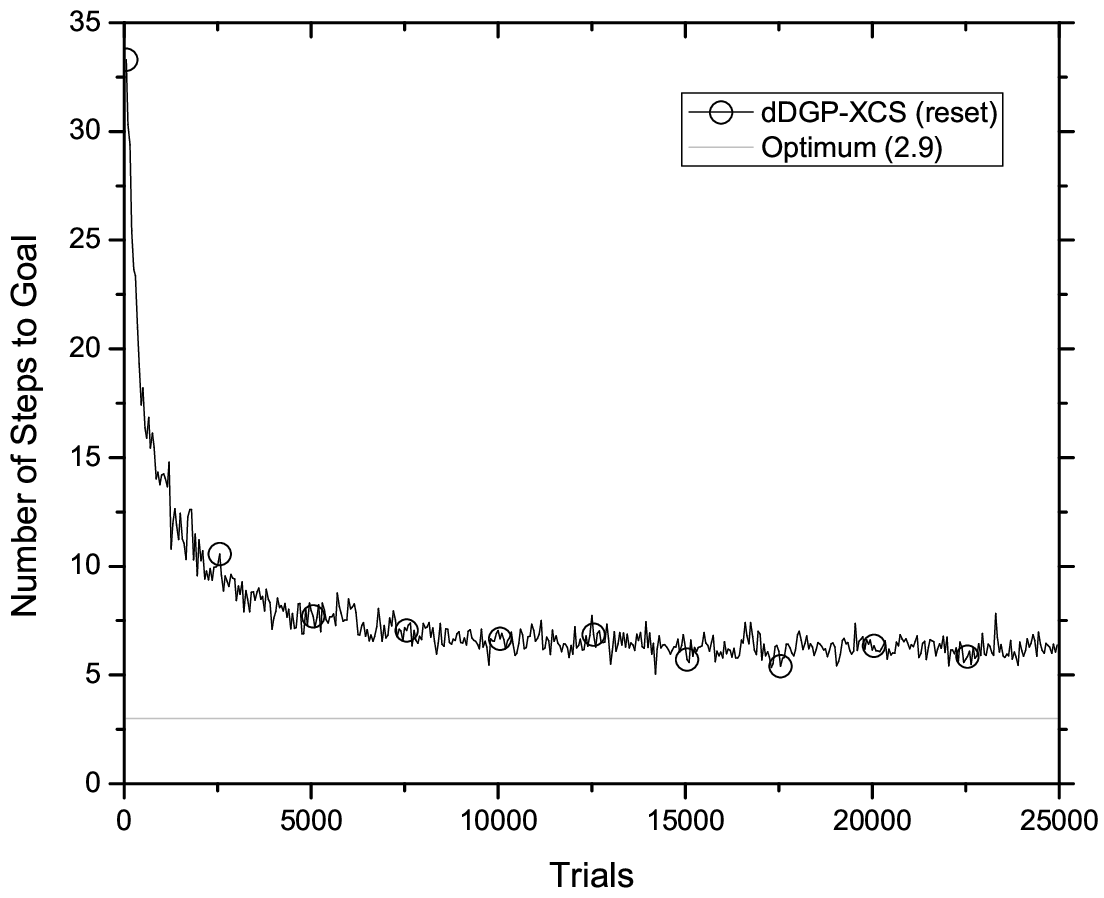,width=\figwidth}
	\caption{dDGP-XCS Woods~101: Number of Steps to Goal with Nodes Reset (circle).}
	\label{fig:dDGP-XCS-Woods101-reset} 
\end{figure}

Figures~\ref{fig:dDGP-XCS-Woods101}--\ref{fig:dDGP-XCS-Woods101-Topology} show the performance in the Woods~101 environment where all parameters used are identical to those applied in the previous Maze~4 environment. As can be seen from Figure~\ref{fig:dDGP-XCS-Woods101}, dDGP-XCS, without node resets, is able to achieve optimal performance in Woods~101 after approximately 12,000 trials (this is slower than XCS using an explicit 1-bit memory register ($\approx$7,000 trials, $P=800$)~\cite{LanziWilson:2000}. Figure~\ref{fig:dDGP-XCS-Woods101-SizeMut} shows the mutation rate and macro-classifiers. Figure~\ref{fig:dDGP-XCS-Woods101-Topology} shows the average number of nodes and connections. Optimal performance is unattainable however when the nodes are reset randomly between matching (Figure~\ref{fig:dDGP-XCS-Woods101-reset}), proving that the system is exploiting the potential for memory within asynchronous RBN here. The mechanism works within XCS because rules/RBN experience each input but need not match on each cycle. Hence for the ambiguous states they remain accurate for the payoff received on providing the action but do so having processed the previous input in an appropriate way, {\em potentially without matching}.

\section{Conclusions}

In this paper a form of XCS has been presented with which to design asynchronous random Boolean networks. It has been shown that XCS is able to design ensembles of RBN that collectively solve a computational task under a reinforcement learning scheme. In particular, it has been shown possible to exploit the inherent dynamics of the representation scheme to solve a non-Markov maze, i.e., without extra mechanisms. Current research is exploring the possibilities of DGP as a general representation scheme by which to solve complex problems with LCS.

%
%\bibliographystyle{abbrv}
%\bibliography{references} 
%
\end{document}